\documentclass[sigconf]{aamas}  % do not change this line!

\usepackage{balance}  % do not change this line -- unless you manually balance your last page

\usepackage{graphicx}
\usepackage{amsmath}
\usepackage{booktabs}
\usepackage{amsfonts}
\usepackage{nicefrac}
\usepackage{MnSymbol} 

\DeclareMathOperator*{\argmin}{arg\,min}
\usepackage[ruled]{algorithm2e}

\setcopyright{ifaamas}  % do not change this line!
\acmDOI{}  % do not change this line!
\acmISBN{}  % do not change this line!
\acmConference[AAMAS'19]{Proc.\@ of the 18th International Conference on Autonomous Agents and Multiagent Systems (AAMAS 2019)}{May 13--17, 2019}{Montreal, Canada}{N.~Agmon, M.~E.~Taylor, E.~Elkind, M.~Veloso (eds.)}  % do not change this line!
\acmYear{2019}  % do not change this line!
\copyrightyear{2019}  % do not change this line!
\acmPrice{}  % do not change this line!

\usepackage{balance}
\begin{document}

\title{Agent Embeddings: A Latent Representation for Pole-Balancing Networks}  % put your title here!

% AAMAS: as appropriate, uncomment one subtitle line; see camera ready instructions
%\subtitle{Blue Sky Ideas Track}
%\subtitle{JAAMAS Track}
%\subtitle{Doctoral Mentoring Program}
%\subtitle{Extended Abstract}                               
%\subtitlenote{Please refrain from using subtitle notes}

% replace this with your author block!
\author{Oscar Chang}
\affiliation{%
  \institution{Columbia University}
  \city{New York}
  \state{New York} 
  \postcode{10027}
}
\email{oscar.chang@columbia.edu}
\author{Robert Kwiatkowski}
\affiliation{%
  \institution{Columbia University}
  \city{New York}
  \state{New York} 
  \postcode{10027}
}
\email{robert.kwiatkowski@columbia.edu}
\author{Siyuan Chen}
\affiliation{%
  \institution{Columbia University}
  \city{New York}
  \state{New York} 
  \postcode{10027}
}
\email{sc3618@columbia.edu}
\author{Hod Lipson}
\affiliation{%
  \institution{Columbia University}
  \city{New York}
  \state{New York} 
  \postcode{10027}
}
\email{hod.lipson@columbia.edu}

\begin{abstract}  % put your abstract here!
We show that it is possible to reduce a high-dimensional object like a neural network agent into a low-dimensional vector representation with semantic meaning that we call \textit{agent embeddings}, akin to word or face embeddings. This can be done by collecting examples of existing networks, vectorizing their weights, and then learning a generative model over the weight space in a supervised fashion. We investigate a pole-balancing task, Cart-Pole, as a case study and show that multiple \textit{new} pole-balancing networks can be generated from their agent embeddings without direct access to training data from the Cart-Pole simulator. In general, the learned embedding space is helpful for mapping out the space of solutions for a given task. We observe in the case of Cart-Pole the surprising finding that good agents make different decisions despite learning similar representations, whereas bad agents make similar (bad) decisions while learning dissimilar representations. Linearly interpolating between the latent embeddings for a good agent and a bad agent yields an agent embedding that generates a network with intermediate performance, where the performance can be tuned according to the coefficient of interpolation. Linear extrapolation in the latent space also results in performance boosts, up to a point.
\end{abstract}



  % put your semicolon-separated keywords here!

\maketitle

%%%%%%%%%%%%%%%%%%%%%%%%%%%%%%%%%%%%%%%%%%%%%%%%%%%%%%%%%%%%%%%%%%%%%%%%%%%%%%%%%%%%%%%%%%%%%%%%%%%%%%%%%
%% start of main body of paper
\begin{figure}[ht]
\begin{center}
\includegraphics[width=0.5\textwidth]{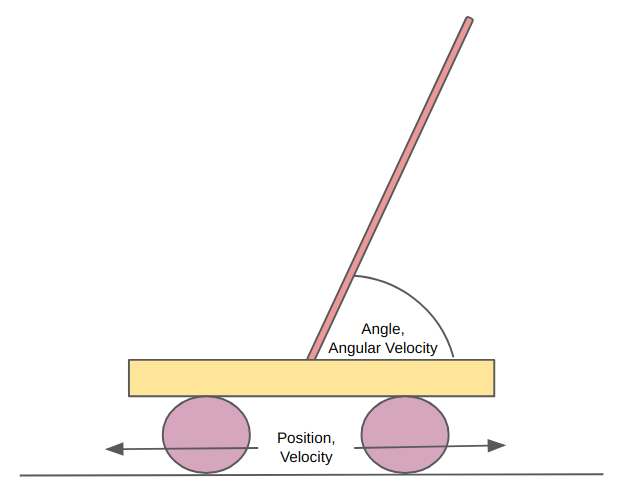}
\end{center}
\caption{Cart-Pole is a game of pole-balancing}
\label{fig:cartpole_diagram}
\end{figure}

\section{Introduction}

Many modern artificially intelligent agents are trained with deep reinforcement learning algorithms \cite{silver2017mastering,mnih2016asynchronous,kulkarni2016hierarchical}. But neural networks have long been criticized for being uninterpretable black boxes that cannot be relied upon in safety-critical applications \cite{zhang2018visual,chakraborty2017interpretability}.

It is important to note, however, that human brains are uninterpretable as well. For example, we know what a face is, because our brains have evolved to detect facial features, and yet, it is nearly impossible to communicate in words what a face is. This problem is especially acute for patients with severe prosopagnosia, who have to rely on other visual cues to identify their friends and family. In fact, it is also quite difficult to communicate precisely the meaning of words. Try talking to a philosopher or a translator about what otherwise ordinary words might mean, \textit{precisely}, and one can be sure to spark a huge debate.

Nonetheless, it is possible to program a computer to detect faces, by reducing high-dimensional images of faces into low-dimensional vector representations with semantic meaning \cite{schroff2015facenet,radford2015unsupervised}. It is also possible to perform sophisticated natural language processing tasks by representing words in a high dimensional vocabulary as low-dimensional vectors \cite{mikolov2013linguistic,pennington2014glove}. Remarkably, these embeddings are amenable to simple linear arithmetic. Take the difference between the latent codes for a face with a mustache and one without a mustache, and one gets something approximating a `mustache' vector. Famously, \citet{mikolov2013linguistic} showed `King' - `Queen' = `Man' - `Woman'.

We propose that a similar strategy can be applied to even something as high-dimensional and complicated as a deep reinforcement learning agent. Our aim is to demonstrate that neural network agents can be compressed into low-dimensional vector representations with semantic meaning, which we term \textit{agent embeddings}. In this paper, we propose to learn agent embeddings by collecting existing examples of neural network agents, vectorizing their weights, and then learning a generative model over the weight space in a supervised fashion.

\subsection{Our Contribution}
As a proof of concept, we report on a series of experiments involving agent embeddings for policy gradient networks that play Cart-Pole, a game of pole-balancing.

We present three interesting findings:
\begin{enumerate}
    \item The embedding space learned by the generative model can be used to answer questions of convergent learning \cite{li2015convergent}, i.e.\ how similar are different neural networks that solve the same task. To our knowledge, we are the first to investigate convergent learning in the context of reinforcement learning agents rather than image classifiers. We extend \citeauthor{li2015convergent}'s work on convergent learning by proposing a new distance metric for measuring convergence between two neural networks. We observe surprisingly that good pole-balancing networks make different decisions despite learning similar representations, whereas bad pole-balancing networks make similar (bad) decisions while learning dissimilar representations.
    \item It has been demonstrated that linear structure between semantic attributes exist in the latent space of a good generative model in the domain of natural language words \cite{mikolov2013linguistic} and faces \cite{radford2015unsupervised}, among other kinds of data. We show that a similar linear structure can be learned in an embedding space for reinforcement learning agents that can be used to directly control the performance of the policy gradient network generated.
    \item We demonstrate that the generative model can be used to recover missing weights in the policy gradient network via a simple and straightforward rejection sampling method. More sophisticated methods of conditional generation are left to future work.
\end{enumerate}

The rest of the paper is organized as follows: we survey the relevant literature (\textit{Related Work}), introduce the pole-balancing task and describe how we learn agent embeddings for it (\textit{Learning Agent Embeddings for Cart-Pole}), present the above-mentioned findings (\textit{Experimental Results and Discussion}), discuss the shortcomings of our approach (\textit{Limitations of Supervised Generation}), speculate on potential applications (\textit{Potential Applications for AI}), and finally summarize the paper at the end (\textit{Conclusion}).

\section{Related Work}
There are four areas of research that are related to our work: interpretability, generative modeling, meta-learning, and Bayesian neural networks.

\subsection{Interpretability}
There has been a lot of recent interest in making reinforcement learning agents and policies interpretable. This is especially important in high-stake domains like health care and education. \citet{verma2018programmatically} proposed to learn policies in a human-readable programming language, while \citet{policycert} proposed to learn certificates that provides guarantees on policy outcomes. \citet{zha2018recognizing} demonstrated utility in learning embeddings for action traces in path planning. \citet{ashlock2013agent}'s work is very similar to ours - they proposed a tool to compare phenotypic differences between solutions found by evolutionary algorithms as a way to explore the geometry of the problem space.

One line of work that has proven useful in increasing our understanding of deep neural network models is that of convergent learning \cite{li2015convergent}, which measures correlations between the weights of different neural networks with the same architecture to determine the similarity of representations learned by these different networks. Convergent learning investigations have hitherto, to our knowledge, only been done on image classifiers, but we extend them to reinforcement learning agents in this paper.

\subsection{Generative Modeling}
Generative modeling is the technique of learning the underlying data distribution of a training set, with the objective of generating new data points similar to those from the training set. Deep neural networks have been used to build generative models for images \cite{radford2015unsupervised}, audio \cite{van2016wavenet}, video \cite{vondrick2016generating}, natural language sentences \cite{bengio2003neural}, DNA sequences \cite{xiao2017dna}, and even protein structures \cite{anand2018generative}. Complex semantic attributes can often be reduced to simple linear vectors and linear arithmetic in the latent spaces of these generative models.

The ultimate (meta) challenge for neural network based generative models is not to generate images or audio, but other neural networks. We use existing networks as meta-training points and use them to train a neural network generator that can produce new pole-balancing networks that do not then need to be further trained with training data from the Cart-Pole simulator. A key advantage of using the same learning framework for both the meta learner and the learner is that this approach could potentially be applied recursively (cue the Singularity).

\subsection{Meta-Learning}
The salient aspect of meta-learning that our work is connected to is the use of neural networks to generate other neural networks. This has been done before in the context of hyperparameter optimization, where one neural network is used to tune the hyperparameters of another neural network \cite{zoph2016neural,pham2018efficient,liu2018darts,smithson2016neural}. \citeauthor{ha2016hypernetworks} proposed the concept of a HyperNet, a neural network that generates the weights of another neural network with a differentiable function. This allows changes in the weights of the generated network to be backpropagated to the HyperNet itself. \citeauthor{chang2018neural} used a neural network to generate its own weights as a way to implement artificial self-replication.

\subsection{Bayesian Neural Networks}
Bayesian neural networks \cite{bishop1997bayesian} maintain a probabilistic model over the weights of a neural network. In this framework, traditional optimization is viewed as finding the maximum likelihood estimate of the probabilistic model. Posterior inference in this case is typically intractable, but variational approximations can be used \cite{kingma2013auto,krueger2017bayesian,louizos2017multiplicative}. Our work involves learning a generative model over the weights of a neural network using existing examples of networks, which is philosophically akin to learning an `empirical Bayesian' prior over the weights in a Bayesian neural network.

\section{Learning Agent Embeddings for Cart-Pole}
\subsection{Supervised Generation}
We propose to learn agent embeddings for neural networks using a two-step process we call \textit{Supervised Generation}. First, we train a collection of neural networks of a fixed architecture to solve a particular task. Next, the weights are saved and used as training input to a generative model. This is a supervised method because we are learning the mapping from a latent distribution to the space of neural network weights by feeding input-output pairs to the model. (There are some obvious downsides to \textit{Supervised Generation} as a method of learning agent embeddings. See the \textit{Limitations of Supervised Generation} section for a detailed discussion.)

In this case, we trained a variational autoencoder (\textit{CartPoleGen}) on the parameter space of a small network (\textit{CartPoleNet}) used to play Cart-Pole.

\subsection{Cart-Pole}
Cart-Pole is a pole balancing task introduced by \citeauthor{barto1983neuronlike} with a modern implementation in the OpenAI Gym \cite{brockman2016openai}. It is also known as the inverted pendulum task and is a classic control problem. The agent chooses to move left or right at every time step with the objective of preventing the pole from falling over for as long as possible. We chose this task because it is easy - around $200$ times easier than MNIST on one measure \cite{li2018measuring} - and hence can be solved with small neural networks.

\subsection{CartPoleNet}
\begin{figure}[ht]
\begin{center}
\includegraphics[width=0.5\textwidth]{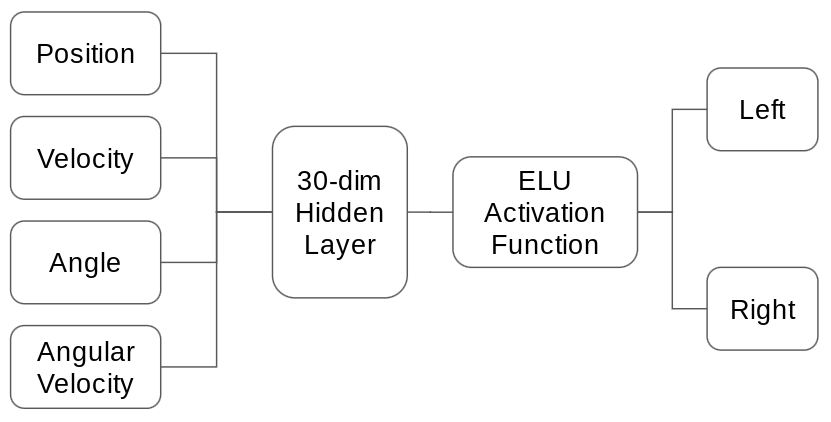}
\end{center}
\caption{Architecture of CartPoleNet}
\label{fig:cartpole_net}
\end{figure}

We devised a simple policy gradient neural network we call \textit{CartPoleNet} with exactly one hidden layer of dimension $30$ (see Figure \ref{fig:cartpole_net}) using the exponential linear unit \cite{clevert2015fast} as the activation function. We collected $74000$ such networks by training them in the Cart-Pole simulator with varying amounts of time, hyperparameters and random seeds for over a week on a cloud computing platform. The $212$-dimensional weight vectors belonging to these $74000$ networks were then used as the training data for the generative model.

A policy gradient neural network approximates the optimal action-value function
\begin{equation}
    Q^{*}(s,a) = \max_\pi \mathbb{E} \left[ \sum_{i=0}^{\infty} \gamma^{i} r_{t+i}  \mid  s_t = s, a_t = a, \pi \right]
\end{equation}
which is the maximum expected sum of rewards $r_i$ discounted by $\gamma$ and achieved by a policy $P(a \mid s)$ that makes an action $a$ after observing state $s$. Cart-Pole assigns a reward of $1$ for every step taken, and each episode terminates whenever the pole angle exceeds $12^\circ$, the position exceeds the edge of the display, or once the pole has been successfully balanced for more than $200$ time steps.

At each epoch, we sample state-action pairs with an epsilon-decreasing policy and store them with their rewards in an experience replay buffer to train the neural network. Note that the neural network only takes state $s$ as input, and its Q-value at action $a$ is represented by the corresponding activation on the last layer. Parametrizing the Q-function with a state-action pair as input is possible but more computationally expensive because it requires $ \mid A \mid $ number of forward passes where $A$ is the action space \cite{mnih2015human}.

\subsection{CartPoleGen}
\begin{figure*}[ht]
\begin{center}
\includegraphics[width=0.95\textwidth]{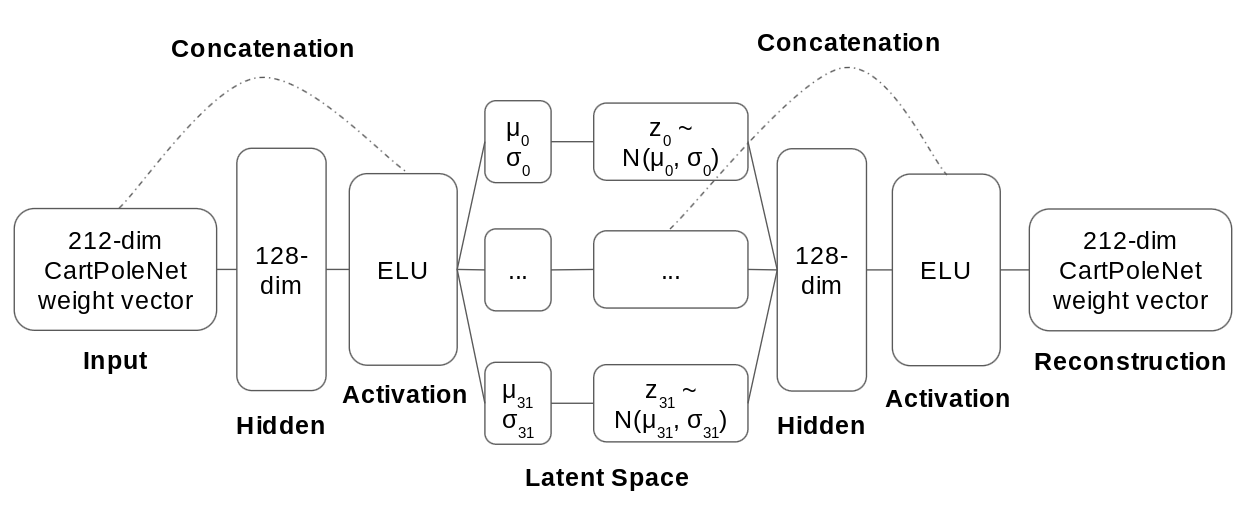}
\end{center}
\caption{Architecture of CartPoleGen}
\label{fig:cartpole_vae}
\end{figure*}

CartPoleGen is a variational autoencoder with a diagonal Gaussian latent space of dimension $32$. It contains skip connections (with concatenation not addition) and uses the exponential linear unit as the activation function as in CartPoleNet (see Figure \ref{fig:cartpole_vae}).

A variational autoencoder \cite{kingma2013auto} is a latent variable model with latent $\mathbf{z}$ and data $\mathbf{x}$. We assume the prior over the latent space to be the spherical Gaussian $p(\mathbf{z}) = \mathcal{N}(\mathbf{z};\mathbf{0},\mathbf{I})$ and the conditional likelihood $p_{\theta}(\mathbf{x} \mid \mathbf{z})$ to be Gaussian, which we compute with a neural network decoder parametrized by $\theta$. The true posterior $p(\mathbf{z} \mid \mathbf{x})$ is intractable in this case, but we assume that it can be approximated by a Gaussian with a diagonal covariance structure that we can compute with a neural network encoder $q_{\phi}(\mathbf{z} \mid \mathbf{x})$ parametrized by $\phi$.

Sampling from the posterior involves reparametrizing $\mathbf{z} \sim \mathcal{N}(\boldsymbol{\mu},\boldsymbol{\sigma})$ to $\mathbf{z} = \boldsymbol{\mu} + \boldsymbol{\sigma} \odot \epsilon$ where $\epsilon \sim \mathcal{N}(\mathbf{0},\mathbf{I})$ to allow the gradients to backpropagate through to $\boldsymbol{\mu}$ and $\boldsymbol{\sigma}$.

We can train the variational autoencoder by maximizing the variational lower bound on the marginal log likelihood of data point $\mathbf{x}$:

\begin{equation}
\begin{aligned}
    \mathcal{L}(\boldsymbol{\theta},\boldsymbol{\phi};\mathbf{x}) = &- \mathcal{D}_{KL}(q_{\phi}(\mathbf{z} \mid \mathbf{x}) \mid  \mid p(\mathbf{z}))\\
    &+ \mathbb{E}_{q_{\phi}(\mathbf{z} \mid \mathbf{x})} \left[ \log p_{\theta}(\mathbf{x} \mid \mathbf{z}) \right]
\end{aligned}
\end{equation}

The Monte Carlo estimator (with latent dimension $k=32$ and noise mini-batch of size $M=1$) for equation (2), also known as the SGVB estimator, becomes

\begin{equation}
\begin{aligned}
    \mathcal{L}(\boldsymbol{\theta},\boldsymbol{\phi};\mathbf{x}) = &\frac{1}{2} \sum_k (1 + \log \sigma_k^2 - \mu_k^2 - \sigma_k^2)\\
    &+ \frac{1}{M} \sum_{m=1}^{M} \log p_{\theta}(\mathbf{x} \mid \mathbf{z}^{(m)})
\end{aligned}
\end{equation}

Notice that maximizing the above lower bound involves maximizing the model's log-likelihood, which is equivalent to minimizing its negative log-likelihood. Minimizing the negative log-likelihood of a Gaussian model is equivalent to minimizing the mean squared error, which is simply the reconstruction cost in an autoencoder.

\subsection{Sampling from CartPoleGen}
We divided the $74000$ networks into four groups depending on the network's \textit{survival time}, which we measure as the average number of steps before the episode terminates across $100$ random testing episodes. The survival time is quite a robust measure of CartPoleNet's performance; it varies $\pm5$ at most due to the stochasticity of the Cart-Pole simulator.

We trained CartPoleGen in two settings. The first setting involves training on all $74000$ networks, and then measuring the survival time of $200$ new samples drawn from the posterior distribution of the variational autoencoder. The second setting involves training a separate CartPoleGen conditioned on each group with a conditional VAE setup \cite{sohn2015learning}. The survival time in the second setting is also measured with $200$ new samples drawn from the posterior of the conditional generative model.

The training was conducted using ADAM \cite{kingma2014adam} for $20$ epochs with a batch size of $10$. The results are summarized in Table \ref{table:cartpole_results}. For comparison, an agent that randomly selects actions lasts on average $22$ steps, and an agent that makes the same action at every time step lasts only $9$ steps. The Cart-Pole simulation ends once an agent has survived $200$ steps, so it is not possible to survive longer than that.

\begin{figure}[ht]
\begin{center}
\includegraphics[width=0.5\textwidth]{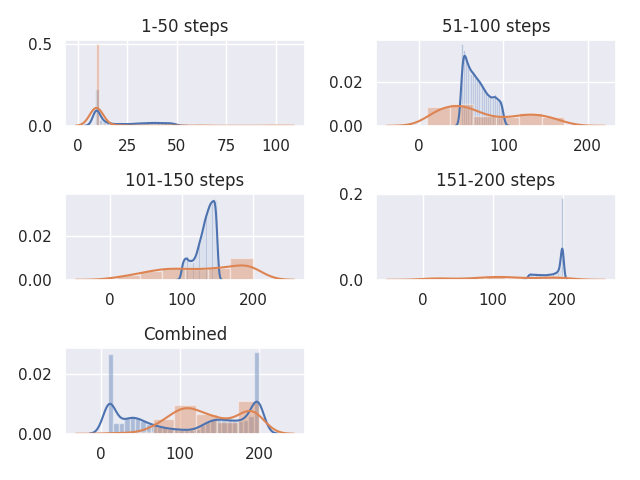}
\end{center}
\caption{The figures are plotted as histograms, with KDE curves fitted on them. The x-axis denotes the survival time, and the y-axis denotes the percentage of networks with that survival time. The figures in blue represent the networks from the trainset, while the figures in orange represent the sampled networks.}
\label{fig:cartpole_kde}
\end{figure}

Figure \ref{fig:cartpole_kde} shows that the CartPoleGen does not accurately capture the exact distribution of the training data, but that it does offer an approximation to it. Training on better networks tends to lead to better generated networks, with the exception of the $151-200$ survival time group. We surmise that this is a consequence of the unimodal variational approximation.

Curiously, CartPoleGen seems to display zero-avoiding rather than zero-forcing behavior, which show that the behavioral properties of neural network agents do not directly match their weight space properties. It is interesting that in some cases, we are able to sample new networks that dramatically outperform the original networks that were in the training set. In the conditional groups, the generated samples typically display much higher variance than is found in the training set, but this does not hold true in the combined setting.

We hypothesize that the approximation gap is partially due to the limitations of the variational autoencoder and can be narrowed with a more expressive generative model. We experimented with various other neural architectures for the encoder and decoder, but did not manage to find significant improvements. In fact, the architecture of CartPoleGen presented here approximates a similar distribution when the encoder and decoder are trained with linear layers.

We also experimented with using GANs \cite{goodfellow2014generative,radford2015unsupervised} as the generative model for CartPoleGen, but did not manage to successfully train them. In our experiments, the discriminator was not able to provide a good teaching signal to the generator because it managed to rapidly distinguish between the fake and real samples.

\begin{table}[ht]
\caption{Sampling new instances of CartPoleNet}
\label{table:cartpole_results}
\centering
\begin{tabular}{ p{2.1cm} p{1.1cm} p{1.77cm} p{1.77cm}  }
 \toprule
 Group & Trainset Size & (Mean, Std) of Survival Time in Trainset & (Mean, Std) of Survival Time in Generated Samples\\
 \midrule
 $1-50$ steps & 25608 & 21.8, 11.5 & 11.0, 9.7\\
 $51-100$ steps & 9400 & 69.7, 14.2 & 77.3, 46.5\\
 $101-150$ steps & 10103 & 132.6, 13.1 & 127.0, 55.3\\
 $151-200$ steps & 28889 & 184.9, 16.3 & 116.4, 58.6\\
 Combined & 74000 & 106.7, 73.3 & 136.7, 42.8\\
 \bottomrule
\end{tabular}
\end{table}

\section{Experimental Results and Discussion}
In this section, we perform three experiments using the agent embeddings learned by CartPoleGen in the previous section. These experiments involve (1) deciding if different CartPoleNets of similar ability learn similar representations, (2) exploring the latent space learned by CartPoleGen, and (3) repairing missing weights in a CartPoleNet.

\subsection{Convergent Learning}
\citeauthor{li2015convergent} posed the question of convergent learning: do different neural networks learn the same representations? In the case of convolutional neural networks used as image classifiers, they found that shallow representations that resemble Gabor-like edge detectors are reliably learned, while more semantic representations sometimes differ.

Success is usually not an accident. Prima facie, for a given complex task, it seems like there can be a million ways to fail it, but only a handful of ways to successfully solve it. We hypothesize this to be the case for Cart-Pole, but found surprisingly that the reverse was true.

\citeauthor{li2015convergent} measured activations on a reference set of images from the ImageNet Large Scale Visual Recognition Challenge 2012 dataset \cite{russakovsky2015imagenet}, and calculated the correlation of such activations between pairs of convolutional neural networks. For CartPoleNets, the inputs are environment states in Cart-Pole, so we had to first collect a reference set of $10000$ diverse states in the Cart-Pole simulator before computing CartPoleNet activations on them.

We follow the same methodology as \citeauthor{li2015convergent} with the slight modification that we use the absolute value of the activations. This is because we use ELUs in CartPoleNet which have important negative activations that ReLU-based networks do not.

\begin{equation}
    \text{Mean}: \mu_i = \mathbb{E}[ |X_i| ]
\end{equation}

\begin{equation}
    \text{Std}: \sigma_i = \sqrt{\mathbb{E}[( |X_i| - \mu_i)^2]}
\end{equation}

\begin{equation} \label{corr_eqn}
     \text{Corr}: \rho_{i, j} = \mathbb{E}[( |X_i| -\mu_i)( |X_j|  - \mu_j)] / \sigma_i \sigma_j
\end{equation}

The correlation between activations of a pair of networks can then be used to pair units from the first network with units from the second. In a bipartite matching, we assign each pair by matching units with the highest correlation, taking them out of consideration, and repeating the process until all the units have been paired. Hence, each unit belongs to exactly one pair. This can be done efficiently with the Hopcroft-Kraft algorithm \cite{hopcroft1973n}. In a semi-matching, we sequentially assign each unit $i$ from the first network using the unit $j$ from the second network with the highest correlation $\rho_{i, j}$. It is thus possible that some units will belong to multiple pairs, while others will not get paired at all.

Two networks are in some sense equivalent if we can arrive at one network by permuting the ordering of the units of the other. The \textit{convergence distance (CD)} between two networks can hence be quantitatively measured as the distance between the bipartite matching and the semi-matching (see Equation \ref{net_dist}). There is exactly one bipartite matching of maximum cardinality, but multiple possible semi-matchings depending on the order of assignment. We compute the convergence distance using the \textit{canonical} semi-matching, defined as the semi-matching performed in descending order from the most highly correlated to the least highly correlated pair in the bipartite matching.

\begin{equation}
\label{net_dist}
    \text{CD}(\text{Net1}, \text{Net2}) = \sum_i \rho_{i, \text{Bipartite}(i)} - \rho_{i, \text{Semi}(i)}
\end{equation}

\begin{figure}[ht]
\begin{center}
\includegraphics[width=0.5\textwidth]{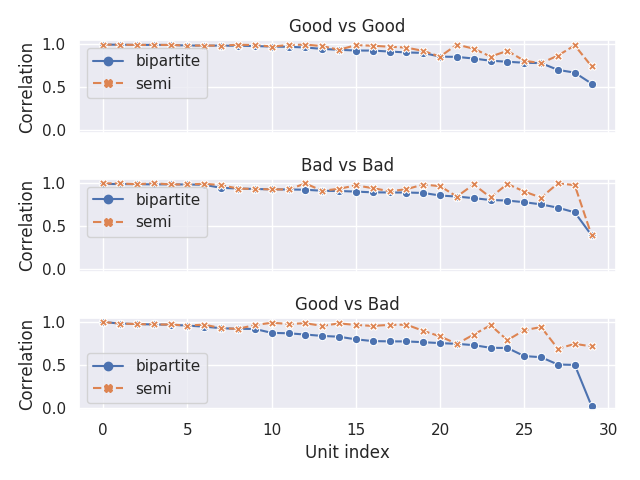}
\end{center}
\caption{The figure shows correlations between \textit{hidden} activations of a pair of good CartPoleNets, a pair of bad CartPoleNets, and a pair with one good and one bad CartPoleNet. For the networks used in this figure, the convergence distances between the pairs are $1.51$, $1.75$ and $3.91$ respectively.}
\label{fig:hidden_convergence}
\end{figure}

\begin{figure}[ht]
\begin{center}
\includegraphics[width=0.5\textwidth]{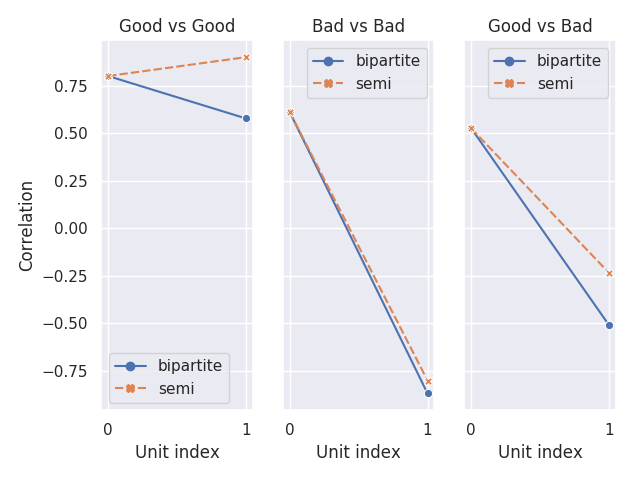}
\end{center}
\caption{The figure shows correlations between \textit{output} activations of a pair of good CartPoleNets, a pair of bad CartPoleNets, and a pair with one good and one bad CartPoleNet. For the networks used in this figure, the convergence distances between the pairs are $0.32$, $0.07$ and $0.28$ respectively.}
\label{fig:output_convergence}
\end{figure}

We sampled ten networks with survival time $\sim$$191$ (from the conditional CartPoleGen trained on the $151$-$200$ survival time group) and ten networks with survival time $\sim$$29$ (from the conditional CartPoleGen trained on the $0$-$50$ survival time group) to represent good and bad networks respectively. Randomly selecting actions results in a survival time of $22$, so $29$ represents a bad network that is nonetheless acting better than random. The average all-pairs convergence distance in the good group and in the bad group are then computed, with the results summarized in Table \ref{table:convergence_results}. We visualize the convergence distances in the hidden and output layer between selected pairs of CartPoleNets in Figures \ref{fig:hidden_convergence} and \ref{fig:output_convergence} respectively.

% good output
% ('mean d', 0.3163136083839668)
% ('std d', 0.48608253507337085)
% ('total_step_list', [190.98, 190.99, 191.03, 191.08, 191.11, 191.19, 191.22, 191.26, 191.28, 191.3])
% bad output
% ('mean d', 0.08665445840193166)
% ('std d', 0.11297030247176831)
% ('total_step_list', [28.75, 29.12, 29.15, 29.15, 29.25, 29.41, 29.73, 29.74, 29.74, 29.86])
% good hidden
% ('mean d', 2.751283103196571)
% ('std d', 1.9554937346011225)
% ('total_step_list', [190.98, 190.99, 191.03, 191.08, 191.11, 191.19, 191.22, 191.26, 191.28, 191.3])
% bad hidden
% ('mean d', 3.1287888503197894)
% ('std d', 1.6882405031708756)
% ('total_step_list', [28.75, 29.12, 29.15, 29.15, 29.25, 29.41, 29.73, 29.74, 29.74, 29.86])

\begin{table}[ht]
\caption{Convergence of Good vs. Bad Networks}
\label{table:convergence_results}
\centering
\begin{tabular}{ p{1.7cm} p{1.5cm} p{1.65cm} p{1.65cm} }
 \toprule
 Group & Survival Time & Mean, Std CD (Hidden) & Mean, Std CD (Output)\\
 \midrule
 Good & 191 & 2.75, 1.96 & \textbf{0.32}, 0.49\\
 Bad & 29 & \textbf{3.13}, 1.7 & 0.09, 0.11\\
 \bottomrule
 \tabularnewline
\end{tabular}
Higher CDs correspond to divergence, while lower CDs correspond to convergence.
\end{table}

The data suggests that for the task of Cart-Pole that there are more ways to be successful than to be bad. In other words, given a random state in the environment, the good networks can diverge in their decision to move left or right to balance the pole, but the bad networks uniformly make the wrong decision. Surprisingly also, despite the good networks displaying divergence in their actions, they pick up on more convergent (good) representations.

It is quite interesting that there are more ways to balance a pole successfully than poorly, but the skills needed for the different paths to success are similar. We hypothesize that this is because the order of actions might be less important than the overall composition of the two actions. Consider a sequence of four actions. \textit{\{Left, Right, Left, Right\}} would be highly negatively correlated with \textit{\{Right, Left, Right, Left\}} but on average, they might produce the same outcome of keeping the pole balanced. On the other hand, \textit{\{Left, Left, Left, Left\}} is highly correlated with \textit{\{Left, Left, Left, Left\}} and they both cause the pole to quickly lose its balance.

\subsection{Exploring the Latent Space}
\begin{figure}[ht]
\begin{center}
\includegraphics[width=0.5\textwidth]{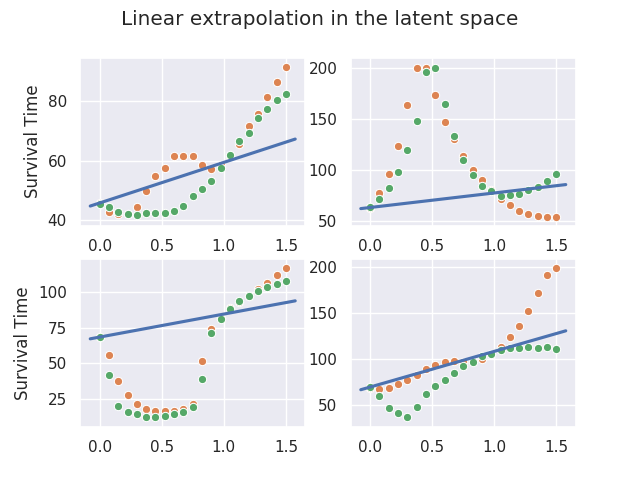}
\end{center}
\caption{The x axis represents the coefficient of interpolation $\alpha$, while the y axis represents the survival time of the sampled networks. The orange dots represent networks sampled from interpolating within the latent space, while the green dots represent networks interpolated within the weight space with the same coefficient of interpolation. The blue line is a straight line drawn from the survival time of the network sampled from the first agent embedding to the survival time of the network sampled from the second agent embedding.}
\label{fig:cartpole_interpolation}
\end{figure}

The latent space in CartPoleGen gives us semantic information about the kinds of networks that can be generated. We selected pairs of agent embeddings and sampled $20$ new embeddings from $\alpha = 0.0$ to $\alpha = 1.5$ where $\alpha$ represents the coefficient of linear interpolation between the pair of embeddings. $0 < \alpha < 1$ represents interpolation, while $\alpha > 1$ represents extrapolation. The results are summarized in Figure \ref{fig:cartpole_interpolation}.

The top left graph represents a pair of agent embeddings with a hidden CD of $1.82$, the top right $12.5$, the bottom left $2.13$, and the bottom right $2.77$. We observe that linearly interpolating within the latent space of CartPoleGen is not the same as simply interpolating within the weight space of CartPoleNet, given that CartPoleGen is non-linear in nature. In many cases, moving from a worse agent embedding to a better one tracks a similar improvement in survival time, as is the case in the top left and bottom right graphs. Furthermore, extrapolation results in a performance boost, up to a point.

However, we also observed many cases where interpolation resulted in agent embeddings whose network performed far worse or far better than the two embeddings used as endpoints for the interpolation. Interestingly, when the interpolated embeddings performed far better, it is often the case that the hidden CDs of the networks used for the two endpoint embeddings is fairly large. In the case of the top right graph, the hidden CD is in fact a few standard deviations above the mean.

\subsection{Repairing Missing Weights}
The generative model can be used to repair CartPoleNets with missing weights. We propose a simple rejection sampling based method (see Algorithm \ref{alg_repair}) to continuously sample new CartPoleNets from the model until suitable candidates are found to fill out the missing weights. We experiment with two possible criteria that can be used to pick the candidate.

\begin{equation}
    W = \text{Existing} \cupdot \text{Missing}
\end{equation}
\begin{equation}
    C = \text{Candidate}
\end{equation}

The \textit{Missing Criterion} (see Equation \ref{eqn:missing}) picks out the candidate who is most similar to the damaged CartPoleNet when we are only comparing the existing weights.

\begin{equation}
\label{eqn:missing}
    C^* = \argmin_C \sum_{i \in \text{Existing}} (W_i - C_i)^2
\end{equation}

The \textit{Whole Criterion} (see Equation \ref{eqn:whole}) picks out the candidate who is most similar to the damaged CartPoleNet. This biases the selection towards finding candidates with tiny weights in the missing space.

\begin{equation}
\label{eqn:whole}
    C^* = \argmin_C \sum_{i \in W} (W_i - C_i)^2
\end{equation}

\begin{algorithm}[ht]
\SetAlgoLined
 $\gamma = 200$\\
 $k = 10$\\
 $\varepsilon = 5$\\
 Sample $\gamma$ networks from CartPoleGen\\
 Pick $k$ best candidates $C^*$ using a \textit{Criterion}\\
 \For{$i \in [k]$}{
    \If{$\mid ST(C_i) - ST(W) \mid\ < \varepsilon$}{\Return Success, $C_i$}
 }
 \Return Failure, $\emptyset$
 \caption{Rejection sampling based method to repair missing weights in a CartPoleNet W. Let $ST()$ represent the survival time of a network.}
 \label{alg_repair}
\end{algorithm}

\begin{figure}[ht]
\begin{center}
\includegraphics[width=0.5\textwidth]{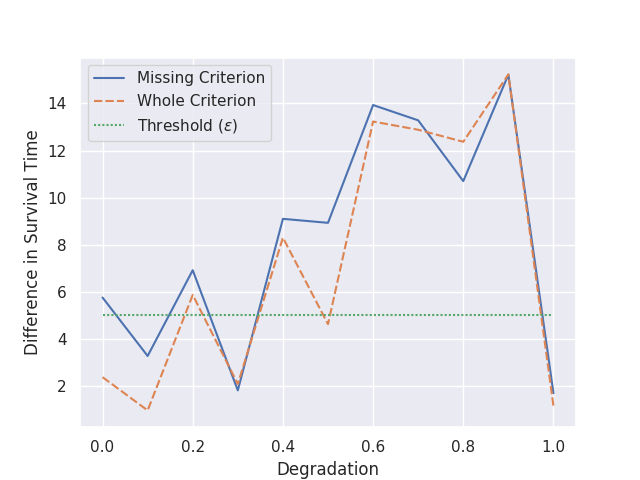}
\end{center}
\caption{The figure shows the performance of the two criteria (in terms of the difference in survival time between the original network and the recovered network) used to repair missing weights at ten different levels of degradation. The threshold $\varepsilon$ represents what we consider a successful level of recovery, so all the points below the threshold represent successful reversal of degradation.}
\label{fig:repair}
\end{figure}

We can probe the limits of our generative model for the task of weight repair by determining how much degradation can be reversed with a fixed computational budget (i.e. $\gamma$ and $k$ are fixed). To investigate this, we fix a given CartPoleNet, degrade it at a fixed level (i.e. zero out a fixed fraction of the weights at random), and repair it using the rejection sampling based algorithm proposed. The results are summarized in Figure \ref{fig:repair}.

We observe that the two criteria seem to perform similarly, with \textit{Whole Criterion} performing slightly better, and we managed to successfully recover the network at some levels of degradation. While we do not recover the network completely (below the acceptable threshold of $5$) in many cases, it is hopeful to note that there is partial recovery (the difference in survival times is at most $15$). It is also interesting that it is possible to recover the network at complete degradation; this suggests perhaps that CartPoleGen has memorized this network.

The scheme described here can also be straightforwardly applied to the task of repairing (or verifying) corrupted weights instead of missing weights. We note that rejection sampling is an inefficient method of doing weight repair, and more sophisticated methods of conditional generation should be used if efficiency is of concern.

\section{Limitations of Supervised Generation}
We note three main limitations of the \textit{Supervised Generation} method in learning agent embeddings.

\subsection{High Sample Complexity}
\begin{figure}[ht]
\begin{center}
\includegraphics[width=0.5\textwidth]{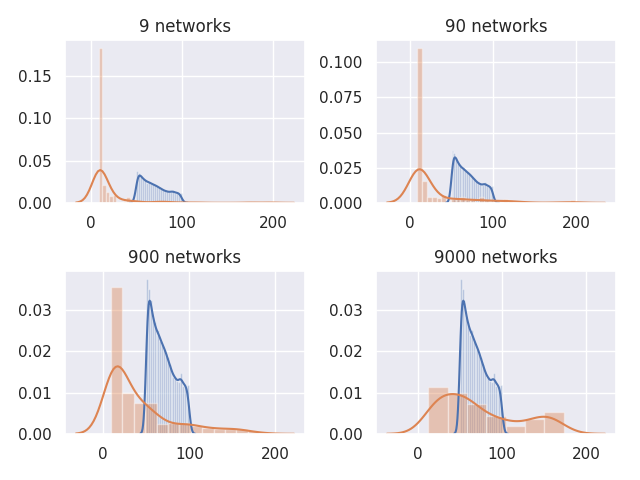}
\end{center}
\caption{If we try to train the distribution in the 51-100 survival time group referred to in Figure \ref{fig:cartpole_kde} with fewer number of samples, we get worse approximations. 
}
\label{fig:cartpole_kde_efficiency}
\end{figure}

One of the primary drawbacks of the \textit{Supervised Generation} method is the two-step process needed to first collect the data then train a generative model on it. This requires training a very large number of networks to provide the generative model with data. Figure \ref{fig:cartpole_kde_efficiency} shows progressively worse approximations when we decrease the number of sampled networks by an order of magnitude.

In principle, an agent embedding does not have to be learned in this manner. For example, it might be possible to do \textit{Online Generation} where a generative model learns to generate new networks on-the-fly with an online algorithm. \textit{Online Generation} will probably be more sample efficient.

\subsection{Subpar Model Performance}
CartPoleGen does not approximate the training distribution very well (see Figure \ref{fig:cartpole_kde}). This might potentially be fixed with a better generative model that also has access to online training data. For example, Bayesian HyperNetworks \cite{krueger2017bayesian} might be a promising candidate.

\subsection{Scaling Issues}
We tried using a variational autoencoder to learn a $21840$-dimensional weight vector for a small neural network that does MNIST image classification. Reinforcement learning agents that process images with CNNs would most likely contain weights at this order of magnitude at minimum. We trained it on a dataset of $10000$ networks each with $>$$95\%$ accuracy, but none of the sampled networks managed to perform with $>$$30\%$ accuracy on a test set.

It might be difficult to scale the \textit{Supervised Generation} method to large networks, even with significant advances made in generative modeling techniques. This is because even state of the art supervised generative models typically deal with data of much lower dimensions ($<$$1000$). A notable exception is WaveNet \cite{van2016wavenet}, but it deals with audio data which is relatively smooth and can tolerate high amounts of error, while the weights of a neural network are very discontinuous and are not robust to small amounts of additive noise.

\section{Potential Applications for AI}
The ultimate challenge for neural network based generative systems is not generating images, sounds, or videos. The ultimate challenge is the generation of other neural networks. Learning agent embeddings is therefore a very difficult goal to accomplish, but we outline several potential applications for AI in general.
\begin{itemize}
    \item AI systems powered by neural networks are often criticized for being uninterpretable. Agent embeddings provide us with a tool to gain insight into its internal workings and the space of possible solutions, which we have demonstrated with the task of pole balancing in this paper.
    \item The generative model can be conditioned to prevent it from generating networks that have undesirable properties like biases or security vulnerabilities. This is helpful for improving the fairness and security of AI systems. We showed how CartPoleGen can be used to repair weights in a network for example, which increases the data integrity of the system.
    \item It is helpful for an AI system to be able to generate worker AIs in a modular fashion. Each worker AI can be represented with its own agent embedding, and the generative model can be a factory that delivers a custom solution conditioned on the task given.
    \item Reinforcement learning agents perform better when they have access to a model of their environment. We think they will also perform better in multi-agent systems when they have access to compressed embeddings of other agents.
\end{itemize}

\section{Conclusion}
In this paper, we presented the concept of agent embeddings, a way to reduce a reinforcement learning agent into a small, meaningful vector representation. As a proof of concept, we trained an autoencoder neural network CartPoleGen on a large number of policy gradient neural networks collected to solve the pole-balancing task Cart-Pole. We showcased three interesting experimental findings with CartPoleGen and described the challenges of the \textit{Supervised Generation} method.

\section{Acknowledgments}
This research was supported in part by the US Defense Advanced Research Project Agency (DARPA) \textit{Lifelong Learning Machines} Program, grant HR0011-18-2-0020. We would like to thank Peter Duraliev for his helpful suggestions in editing the paper.

%%%%%%%%%%%%%%%%%%%%%%%%%%%%%%%%%%%%%%%%%%%%%%%%%%%%%%%%%%%%%%%%%%%%%%%%%%%%%%%%%%%%%%%%%%%%%%%%%%%%%%%%%
%% bibliography: see CFP for number of permitted pages

\bibliographystyle{ACM-Reference-Format}  % do not change this line!
\balance  % do not change this line -- unless you manually balance your last page
\bibliography{aamas}  % put name of your .bib file here

\end{document}